\pdfoutput=1

\documentclass[11pt]{article}

\usepackage[]{emnlp2021}

\usepackage{times}
\usepackage{latexsym}

\usepackage[T1]{fontenc}

\usepackage[utf8]{inputenc}

\usepackage{microtype}

\usepackage{multirow}
\usepackage{booktabs}
\usepackage{amsmath}
\usepackage{amssymb}
\usepackage{graphicx}

\newcommand{\bulletc}{\noindent$\bullet\;$}

\DeclareMathOperator{\argmax}{argmax}

\title{Towards Building ASR Systems for the Next Billion Users}

\author{
    Tahir Javed$^{1,2}$ \hspace{0.2cm} Sumanth Doddapaneni$^{2,4}$ \hspace{0.2cm} Abhigyan Raman$^{2}$ \\
    \textbf{Kaushal Santosh Bhogale}$^{2}$ \hspace{0.2cm} \textbf{Gowtham Ramesh}$^{2,4}$ \hspace{0.2cm} \textbf{Anoop Kunchukuttan}$^{2,3}$ \\
    \textbf{Pratyush Kumar}$^{2,3}$ \hspace{0.2cm}  \textbf{Mitesh M. Khapra}$^{1,2,4}\thanks{Corresponding author: miteshk@cse.iitm.ac.in}$
    \\ \\
    $^1$IIT Madras,
    $^2$AI4Bharat,
    $^3$Microsoft,
    $^4$RBCDSAI
}

\begin{document}

\maketitle

\begin{abstract}
Recent methods in speech and language technology pretrain very LARGE models which are fine-tuned for specific tasks. However, the benefits of such LARGE models are often limited to a few resource rich languages of the world. In this work, we make multiple contributions towards building ASR systems for low resource languages from the Indian subcontinent. First, we curate 17,000 hours of raw speech data for 40 Indian languages from a wide variety of domains including education, news, technology, and finance. Second, using this raw speech data we pretrain several variants of wav2vec style models for 40 Indian languages. Third, we analyze the pretrained models to find key features: codebook vectors of similar sounding phonemes are shared across languages, representations across layers are discriminative of the language family, and attention heads often pay attention within small local windows. Fourth, we fine-tune this model for downstream ASR for 9 languages and obtain state-of-the-art results on 3 public datasets, including on very low-resource languages such as Sinhala and Nepali. Our work establishes that multilingual pretraining is an effective strategy for building ASR systems for the linguistically diverse speakers of the Indian subcontinent. Our code, data and models are available publicly at \href{https://indicnlp.ai4bharat.org/indicwav2vec/}{IndicWav2Vec} and we hope they will help advance research in ASR for Indic languages.
\end{abstract}

\section{Introduction}

The Indian subcontinent is one of the most linguistically diverse regions in the world as well as one of the most populous regions in the world - little wonder that it is called a `subcontinent'. It is home to around 2 billion people from 7 countries who speak 100+ major languages (and thousands of minority languages and dialects) belonging to four major language families\footnote{https://en.wikipedia.org/wiki/Languages\_of\_South\_Asia}$^,$\footnote{https://en.wikipedia.org/wiki/Languages\_of\_India}. Of these, the Indo-Aryan and Dravidian languages are spoken by a large section of the population. These language families have intermingled over a large period of time giving rise to the Indian linguistic area/sprachbund where languages across these families share many features \cite{emeneau1956india,subbarao2012south,kunchukuttan2020utilizing}. 

Building high-quality ASR models for such a large and diverse pool of languages is challenging, even if we restrict ourselves to 30 odd languages which have more than a million speakers. Many modern ASR models rely on large amounts of labeled data for each language to build high-quality ASR systems. Such approaches are expensive and not scalable, thus limiting the reach of ASR technologies to some languages and a section of the population. 
In addition to these challenges on availability of labeled data, Indian languages also face a set of common challenges that need to be addressed. Most Indic scripts have a larger character set than English. The complex inflectional systems of Indian languages make modelling phonotactics more challenging. The rich inflectional/agglutinative nature of Indian languages results in larger vocabulary sizes, presenting challenges to incorporating language models. At the same time there are opportunities from a unified perspective. Collection of unlabeled data for pretraining can be undertaken as a joint effort since many sources might be shared amongst these languages. A largely overlapping phoneme inventory, logically overlapping character sets and syntactic similarity can help utilize linguistic similarity at various levels to build multilingual models where transfer learning can be effective.

In this context, it is important to take note of recent work that has demonstrated the benefits of unsupervised pretraining and multilingual fine-tuning to significantly improve ASR quality for low-resource languages \cite{baevski2020wav2vec,conneau2020unsupervised}. 
In particular, the wav2vec model has established two key results for English ASR. 
One, an end-to-end DNN architecture borrowing the popular Transformer architecture from NLP establishes SOTA results.
And two, pretraining on a large corpus of data reduces the requirement for labeled fine-tuning from hundreds of hours to few hours and to even tens of minutes.
It is worthwhile to explore if these findings from English ASR transfer to Indic ASR, especially given the diversity and above mentioned challenges with Indic languages.
More specifically, would a wav2vec-like model establish SOTA results on available benchmarks across Indic languages?
Further, would large pretraining preclude the need for collecting large amounts of labeled data?
And finally, would pretraining a multilingual model across Indian languages provide positive benefits across these related languages?
In order to answer these questions, we make the following contributions:

\bulletc We curate 17,314 hours of raw audio data for pretraining across 40 languages from 4 language families, making it one of the largest and most diverse collections of Indic language data. This data has been curated in a short time frame from public sources which have a permissive license. It indicates that the feasibility of collecting large amount of pretraining data and further efforts can be made to significantly expand this collection. 

\bulletc Starting from the wav2vec 2.0 model, we perform extensive ablation studies on architecture choices, pretraining and fine-tuning strategies, language models and choice of lexicon to arrive at a training and decoding regimen that works well for Indian languages.

\bulletc Our ASR models achieve SOTA performance on 9 Indian languages on 3 publicly available benchmarks with small fine-tuning datasets. These results indicate that end-to-end ASR systems based on multilingual pretraining with the wav2vec model hold promise for Indic languages.

\bulletc Our ablation studies reveal that the accuracy of the ASR system on Indic languages sensitively depends on the size of the pretraining corpus, amount of labelled data for fine-tuning, and access to task-specific lexicon.

In summary, we establish that the recent advances of pretraining wav2vec models transfer to Indic ASR and achieve SOTA results against models proposed over multiple years.
However, unlike in the reported results of English ASR, we observe that the WER reported for Indic ASR is significantly higher and sensitively depends on availability of resources: pretraining corpus, fine-tuning data, and task-specific language information.
This suggests that the ASR task on Indic languages remains far from being solved and requires model innovation and continued efforts on curating resources.
We publicly release all the artifacts of our work to spur further work in the area of Indic ASR. This includes: (a) sources of pretraining data along with scripts for their collection and pre-processing, (b) pretraining,  fine-tuning and decoding scripts, (c) language models, and (d) our best ASR models.

\section{Curating speech data for Indian languages}
As mentioned earlier, existing wav2vec models for English \cite{baevski2020wav2vec} and some other languages \cite{conneau2020unsupervised, wang2021voxpopuli} have been trained using 1000s of hours of curated raw audio data in these languages. However, for Indian languages there is no such publicly available repository of audio data. Hence, as our first task, we curate audio data from online sources such as YouTube. 

We hired native speakers in each of the 22 constitutionally recognised Indian languages to help us with this task. These workers were employees of a data collection agency which specializes in creating and curating text and speech data in Indian languages with prior experience of working on similar tasks. The workers were asked to discover channels, playlists or individual videos on YouTube by using a wide variety of keywords from different domains such as education, news, technology, sports, and finance. Once the content was discovered, they were asked to ensure that the audio was clean, \textit{i.e.}, it did not contain any background music and the content was predominantly in the target language. Note that some amount of code-mixing with English is unavoidable in Indian speech content but the workers manually checked that the audio was largely in the target language. The workers were also instructed to ensure speaker diversity by not taking a large amount of content from a single channel. Lastly, the workers were asked to ensure that the video was available under a Creative Commons License. This was important to ensure that we can freely share the URLs with the research community, ensuring reproducibility of our work, although this significantly limited the amount of data we could collect, especially in the low resource Indian languages. For each video which cleared the above checks, the workers were asked to copy the URL in an excel sheet and mention the target language.

Apart from YouTube, we also curated content from  \texttt{newsonair}\footnote{http://newsonair.com/} which is a radio news channel run by the Govt.~of India and broadcasts news in multiple Indian languages. We did not need any of the above manual checks for \texttt{newsonair} as being a large, professional news channel, the content there was already clean (e.g., the language of the audio was clearly mentioned and the audio files were largely free of any background music). We crawled their website to get the URLs of audio files in 40 Indian languages.

\begin{table}[!t]
\small
\begin{tabular}{lrr||lrrr}
\toprule
\textbf{Language} & \textbf{NoA} & \textbf{YT}  & \textbf{Language} & \textbf{NoA} & \textbf{YT}  \\
\midrule
\textbf{assamese} & 667 & 176  & \textbf{ladakhi} & 121 & 0  \\
\textbf{balti} & 7 & 0  & \textbf{lepcha} & 52 & 0  \\
\textbf{bengali} & 740 & 295  & \textbf{maithili} & 37 & 1  \\
\textbf{bhojpuri} & 62 & 0  & \textbf{malayalam} & 625 & 232  \\
\textbf{bhutia} & 42 & 0  & \textbf{manipuri} & 374 & 90  \\
\textbf{bodo} & 53 & 11  & \textbf{marathi} & 387 & 667  \\
\textbf{chhatisgar} & 73 & 0  & \textbf{mizo} & 304 & 0  \\
\textbf{dogri} & 606 & 8  & \textbf{nagamese} & 210 & 0  \\
\textbf{garo} & 63 & 0  & \textbf{nepali} & 374 & 333  \\
\textbf{gojri} & 197 & 0  & \textbf{odia} & 716 & 302  \\
\textbf{gujarati} & 375 & 686  & \textbf{pahari} & 24 & 0  \\
\textbf{hindi} & 353 & 722  & \textbf{punjabi} & 562 & 301  \\
\textbf{english} & 408 & 0  & \textbf{rajasthani} & 134 & 0  \\
\textbf{jaintia} & 57 & 0  & \textbf{sambalpuri} & 192 & 0  \\
\textbf{kannada} & 425 & 587  & \textbf{sanskrit} & 152 & 348  \\
\textbf{karbi} & 56 & 0  & \textbf{santali} & 0 & 9  \\
\textbf{kashmiri} & 417 & 19  & \textbf{sindhi} & 77 & 30  \\
\textbf{khasi} & 85 & 0  & \textbf{tamil} & 815 & 197  \\
\textbf{kokborok} & 185 & 0  & \textbf{telugu} & 395 & 657  \\
\textbf{konkani} & 471 & 28  & \textbf{urdu} & 359 & 362 \\
\bottomrule
\end{tabular}
\caption{Number of hours audio data per language after applying the preprocessing steps. Here NoA refers to \texttt{newsonair} data, YT refers to YouTube data.}
\label{table:pretrain-data-table}
\end{table}

Once the URLs were available, we used the youtube-dl library\footnote{https://github.com/tpikonen/youtube-dl} to download the videos from YouTube.  
We then used the FFmpeg library\footnote{https://ffmpeg.org/} to extract the audio data from the videos. For \texttt{newsonair}, these two steps were not needed as we could simply download the audio files from the URLs. We noted that existing speech datasets typically contain mono channel audio data sampled at a sampling frequency of 16Khz. However, in our case, since the data was curated from diverse sources, some of the content was not mono channel and the sampling frequency varied from 8kHz to 44 kHz. To standardise the data, we once again used the FFmpeg library to (i) upsample (downsample) the data which was originally sampled at a frequency lesser (greater) than 16 kHz and (ii) reduce the number of audio channels to 1. We further refined the data by removing long silences in the audio clips using the py-webrtcvad\footnote{https://github.com/wiseman/py-webrtcvad} library which is a python interface to the popular WebRTC VAD (Voice Activity Detection) module developed by Google\footnote{https://webrtc.org/}. The VAD algorithm filters out non-speech content and allows us to set an aggressiveness parameter, which is an integer between 0 and 3 (0 is the least aggressive about filtering out non-speech, 3 is the most aggressive). We found that setting this parameter to 2 worked best for our data. Next, to avoid including highly noisy content in our data, we used Waveform Amplitude Distribution Analysis (WADA-SNR) \cite{kim08e_interspeech} and filtered out audio clips which had a signal-to-noise-ratio (SNR) less than 15 dB. This threshold was chosen after experimenting with a subset of the audio data. Finally, as is standard practice, we chunked audio files so that the maximum duration of any audio file was only 25 secs. 

The final statistics of the data thus obtained in each language are summarised in Table \ref{table:pretrain-data-table}. The uncompressed size of the data is 1.5 TB. The URLs of all the audio files as well as the scripts used for downloading, standardising and cleaning the data will be made publicly available.

\section{IndicWav2Vec: A multilingual ASR model for Indian Languages}
In this section, we describe the methodology followed for pretraining a wav2vec ASR model for Indian languages using the raw audio data described in the previous section. In particular, we describe (i) the model architecture, (ii) the pretraining procedure, (iii) the fine-tuning and decoding procedure and (iv) the procedure for rescoring using an external transformer based language model (LM). Note that, for all of the above, we closely follow the procedure described in \cite{baevski2020wav2vec} for building an English ASR system. 
However, we provide the details here for the sake of completeness and reproducibility.

\subsection{Model architecture}

We use the same architecture as wav2vec 2.0 which consists of (i) a feature encoder for encoding raw audio into a sequence of $T$ latent representations, (ii) a transformer for learning contextualised representations for each of the $T$ units, and (iii) a quantizer for discretizing the representations learnt by the feature encoder to obtain the targets for self-supervised learning. The feature encoder is a multilayered convolutional neural network which takes raw audio signal as input and outputs a sequence of frames $\mathcal{Z} = \{\mathbf{z}_1, \mathbf{z_2}, \dots, \mathbf{z}_T\}$ where $T$ is the number of timesteps and $\mathbf{z}_i \in \mathbb{R}^d$. 

The output $\mathcal{Z}$ of the encoder is fed to a transformer based context network which computes a contextualised representation for each of the $T$ units in the input as $\mathcal{C} = \{\mathbf{c}_1, \mathbf{c}_2, \dots, \mathbf{c}_T\}$, where $\mathbf{c}_i$ is the contextual representation for the $i$-th input (i.e., $\mathbf{z}_i$).

Following standard practice, we augment the $\mathbf{z}_i$'s with positional information before feeding them to the context network using a convolutional layer similar to that used in \cite{baevski2020wav2vec, mohamed2020transformers, baevski2020effectiveness, wu2019pay}. 

The final component of the architecture is a quantizer which takes $\mathcal{Z}$ as input and discretizes it to a finite set of representations using product quantization \cite{5432202}. The output of the quantizer is a sequence of representations  $\mathcal{Q} = \{\mathbf{q}_1, \mathbf{q}_2, \dots, \mathbf{q}_T\}$ where $\mathbf{q}_i$ is the quantized representation for the $i$-th input (i.e., $\mathbf{z}_i$).. 

\subsection{Pretraining objectives and data sampling}

To train the model, we mask a certain fraction of the $T$ input representations before feeding them to the context network. However the inputs to the quantizer are not masked since they serve as the target for the self-supervised objective. This objective is implemented as minimizing a contrastive loss \cite{10.1145/1143844.1143891} that involves maximizing the similarity of the context network's output to the quantized representation for masked input at time $t$, while minimizing the similarity with quantized representations of other masked inputs. In addition to the above contrastive loss, we also use a diversity loss \cite{baevski2020wav2vec} which ensures better utilisation of the codebook.

\noindent\textbf{Data sampling.} While pretraining multilingual models involving a large number of languages with different amounts of training data in each language, it is important to ensure that all languages get a fair representation in the training. In particular, languages which account for a larger share of the training data should not dominate the training. We use temperature based sampling  \cite{DBLP:journals/corr/abs-1907-05019,Conneau2020UnsupervisedCR, pmlr-v139-wang21y} to form multilingual batches \cite{devlin-etal-2019-bert,DBLP:conf/nips/ConneauL19} during pretraining. Let $L$ be the set of all languages used for pretraining. For each language $l \in L$, we draw audio samples from a multinomial distribution $p_l\sim\left(\frac{n_l}{N}\right)^{\alpha}$ where $n_l$ is the number of hours of pretraining data in language $l$, $N$ is the total number of hours of pretraining data across all languages, and $\alpha \in [0, 1]$. Smaller the value of $\alpha$, more is the upsampling for the lower resource languages. 

\subsection{Fine-tuning}
Once the model is pretrained, we fine-tune it for ASR using task-specific supervised data. To do so, we add a randomly initialised projection layer on top of the context network. This layer maps each $d$ dimensional output of the context network, into a $C$ dimensional output where $C$ is the size of the vocabulary. We then use the softmax function to compute a distribution over the vocabulary. In our case, the vocabulary contains all the unique characters in the language. The model can either be jointly fine-tuned using data from all the languages or individually fine-tuned for a specific language. We tried both and for multilingual fine-tuning, we converted the text data into a single script, i.e., the Devanagari script. Though each of these scripts have their own Unicode codepoint range, it is possible to get a 1-1 mapping between characters in these different scripts since the Unicode standard accounts for the similarities between these scripts \cite{kunchukuttan2020indicnlp}. This results in a smaller, compact output vocabulary while enabling better transfer across languages \cite{9414961, DBLP:journals/corr/abs-2104-05596}. We fine-tune the model using the standard CTC loss function \cite{10.1145/1143844.1143891, baevski2020wav2vec, baevski2020effectiveness}. Following \cite{baevski2020wav2vec}, we augment data with a modified version of SpecAugment \cite{park2019specaugment}. 

\subsection{Decoding}
To decode the emissions from the softmax layer, we use a lexicon and a separately trained word-level n-gram language model. Let $\mathbf{y} = \{y_1, y_2, \dots, y_M\}$ be a candidate sequence. Further, let $p_{AM}(\mathbf{y})$ and $p_{LM}(\mathbf{y})$ be the probability assigned to the sequence $\mathbf{y}$ by the fine-tuned network and the language model respectively. We select the optimal word sequence $\mathbf{y^*}$ as follows:
\begin{align}
\label{eq:decoding}
    \mathbf{y^*} = \argmax_{\mathbf{y}} \log p_{AM}(\mathbf{y}) + \alpha \log p_{LM}(\mathbf{y}) + \beta |\mathbf{y}|
\end{align}
where $|\mathbf{y}|$ is the length of the sequence and $\alpha$ and $\beta$ are hyperparameters. We use an efficient beam search decoder \cite{Pratap_2019,liptchinsky2019letterbased} to search candidates while combining network scores, LM scores and word insertion bonuses. 

\subsection{Rescoring}
Note that in the above decoding process, we use an $n$-gram KenLM language model \cite{heafield-2011-kenlm}. However, recent advances in language modeling have shown that transformer based language models perform well. To get the best of both worlds, we optionally use an external transformer based language model to rescore the $n$-best hypothesis produced above \cite{synnaeve2019end}. Specifically, for each output in the $n$-best list, we compute a new score by combining the decoder log-probability, with the weighted log-probability of an external LM, $p_{ELM}(\mathbf{y})$. We then select the best sequence using the following equation:

\begin{equation}
\begin{split}
\label{eq:decoding2}
    \mathbf{y^*_{ext}} =  \argmax_{\mathbf{y}} \log p_{AM}(\mathbf{y}) + \alpha _1\log p_{LM}(\mathbf{y}) \\
    + \alpha _2\log p_{ELM}(\mathbf{y}) + \beta |\mathbf{y}|
\end{split}
\end{equation}
where  $|\mathbf{y}|$ is the length of the sequence and $\alpha _1$, $\alpha _2$ and $\beta$ are hyperparameters.

\section{Experimental Setup}
In this section we describe (i) the datasets used for our experiments, (ii) the hyperparameters used for pretraining and fine-tuning, and (iii) the corpora and hyperparameters used for training the language models.

\noindent\textbf{Pretraining Dataset}
From the 17k hours of raw audio data, we use 3\% data from each language as validation data for tracking the per language validation loss during pretraining.

\noindent\textbf{Fine-tuning Datasets}
We experiment with 3 ASR datasets covering 9 Indian languages. These include the MSR (Microsoft Research) dataset which was released as a part of the \textit{Low Resource Speech Recognition Challenge for Indian Languages} \cite{srivastava18_sltu}, the MUCS2021 dataset which was released as a part of the \textit{Multilingual and code-switching ASR challenges for low resource Indian languages} \cite{diwan2021multilingual}, and a subset of the OpenSLR dataset \cite{kjartansson-etal-sltu2018} obtained from the authors of \citet{9414961}. The train-val-test splits across datasets are summarised in Table \ref{table:ft-dataset}.

\begin{table}[h!]
\resizebox{\linewidth}{!}{
\begin{tabular}{cccc||ccc||ccc}
\toprule
 & \multicolumn{3}{c}{MSR} & \multicolumn{3}{c}{MUCS} & \multicolumn{3}{c}{OpenSLR} \\
 \cmidrule{2-10}
 & Train & Valid & Test & Train & Valid & Test & Train & Valid & Test \\
 \midrule
bn & - & - & - & - & - & - & 70.4 & 4.8 & 4.8 \\
gu & 40.0 & 5.0 & 5.0 & 40.0 & 5.0 & 5.26 & - & - & - \\
hi & - & - & - & 95.05 & 5.55 & 5.49 & - & - & - \\
mr & - & - & - & 93.89 & 5 & 0.67 & - & - & - \\
ne & - & - & - & - & - & - & 70.2 & 5.0 & 5.0 \\
or & - & - & - & 94.54 & 5.49 & 4.66 & - & - & - \\
si & - & - & - & - & - & - & 70.5 & 4.8 & 4.8 \\
ta & 40.0 & 5.0 & 4.2 & 40.0 & 5.0 & 4.41 & - & - & - \\
te & 40.0 & 5.0 & 4.2 & 40.0 & 5.0& 4.39 & - & - & - \\
\bottomrule
\end{tabular}
}
\caption{Size of train-val-test splits across datasets (in hours)}
\label{table:ft-dataset}
\end{table}

\noindent\textbf{Pretraining setup}
We pretrain two variants of the model, \textit{viz.}, BASE and LARGE. Similar to the BASE model in  \citet{baevski2020wav2vec}, our BASE model contains 7 convolutional layers each with 512 channels, strides of (5,2,2,2,2,2,2) and kernel widths of (10,3,3,3,3,2,2). The BASE model has 12 transformer blocks with model dimension 768, and FFN dimension 3072 with 8 attention heads. Similarly, same as \citet{baevski2020wav2vec}, our LARGE model has the same CNN encoder as the BASE model and differs only in the transformer architecture. The LARGE model has 24 transformer blocks with model dimension 1024 and FFN dimension 4096 with 16 attention heads. For both the architectures, we use $G=2$ (codebooks) with $V = 320$ entries per codebook in the quantization module. Instead of pretraining the models from scratch we start with the pretrained checkpoint of the corresponding (base or large) English wav2vec 2.0 model\footnote{https://github.com/pytorch/fairseq/tree/master/examples/wav2vec\#pre-trained-models}. We then further pretrain the model using the data we curated for Indian languages. As mentioned earlier, to account for the skew in the amount of data across different languages, we used temperature based sampling. We experiment with three different values of $\alpha \in \{0.5, 0.7, 1\}$ and found that 0.7 works best in terms of model accuracy in distinguishing the correct masked unit from the distractor units. 

For the BASE model, we crop audio segments to 250k samples or 15.6 seconds of audio. We restrict the max tokens per GPU to 3M and train the model on 24 A100 GPUs with gradient accumulation of 2 steps with the effective batch size of 3.2 hours. We used Adam optimizer with learning rate set to 0.0005 and decayed the learning rate polynomially after a warmup for 32k steps. We trained the model for 160k steps. For the LARGE model we crop audio segments to 320k samples or 20 seconds of audio. We restrict the max tokens per GPU to 1.2M and train the model on 24 A100 GPUs with gradient accumulation of 6 steps making the effective batch size as 3 hours. We again used the Adam optimizer but with learning rate set to 0.005 and decayed the learning rate polynomially without any warmup. We trained the model for 110k steps. We set the random seed to 1 for all experiments.
All other hyper-parameters are set to the default values from the original wav2vec 2.0 code base. Our BASE and LARGE models have 95M and 317M parameters respectively. Our models are implemented in fairseq \cite{ott2019fairseq} and we use fp16\footnote{https://github.com/NVIDIA/apex} operations to speed up training and reduce memory usage. 

\begin{table*}[ht!]
\small\centering
\begin{tabular}{lccc|cccccc|ccc}
\toprule
 & \multicolumn{3}{c|}{MSR} & \multicolumn{6}{c|}{MUCS} & \multicolumn{3}{c}{OpenSLR} \\
 \cmidrule{2-13}
 & gu & ta & te & gu & hi & mr & or & ta & te & bn & ne & si \\
 \midrule
M0: No pretraining & 46.0 & 37.5 & 35.5 & 53.2 & 48.1 & 87.1 & 73.4 & 44.8 & 44.7 & 36.0 & 78.8 & 37.0 \\
M1: IndicW2V$_{b}$ (EkStep data) & 23.4 &	24.0 & 25.8 & 30.3 & 18.0 & 26.5 & 28.7 & 29.9 & 33.2 & 19.7 & 14.4 & 31.0  \\
M2: IndicW2V$_{b}$ (our data) & 22.8 & 23.7 & 24.9 & 29.4 & 17.8 & 24.3 & 27.2 & 29.3 & 31.9 & 18.1 & 13.8 & 24.3  \\
M3: IndicW2V$_{l}$ (our data) & 20.5	& 22.1 &	22.9 &	26.2 &	16.0 &	19.3 &	25.6 &	27.3 &	29.3 &	16.6 & 11.9 &	24.8  \\
M4: $\quad$ + LM$_{small}$ & 16.6 &	14.9 & 14.4 & 18.0 & 16.3 & 14.8 & 19.0 & 25.4 & 22.4 & 14.3 & 13.0 & \textbf{18.6} \\
M5: $\quad$ + LM$_{large}$ & \textbf{11.7} & \textbf{13.6} & \textbf{11.0} & 17.2 & 14.7 & 13.8 & \textbf{17.2} & 25.0 & 20.5 & 13.6 & 13.6 & - \\
M6: $\quad\quad$ + augmented lexicon & 12.3 & 15.1 &	12.4 &	14.8 &	10.5 & 12.2 & 21.9 & 20.0 & 15.2 & 10.6 & 9.7 & -  \\
M7: $\quad\quad\quad$ + Rescoring & 11.9 & 14.8 & 12.0 & \textbf{14.3} & \textbf{9.5} & \textbf{11.7} & 20.6 & \textbf{19.5} & \textbf{15.1} & \textbf{10.5} & \textbf{9.4} & - \\
\bottomrule
\end{tabular}
\caption{Comparison of different choices for pretraining, fine-tuning, and decoding. IndicW2V$_{b}$ and IndicW2V$_{l}$ refer to our base and LARGE models respectively.  LM$_{small}$ refers to the language model trained using transcripts from the training and validation data and LM$_{large}$ refers to the one trained using IndicCorp in addition to the transcripts.}
\label{table:ablation-results}
\end{table*}

\noindent\textbf{Fine-tuning setup}
During fine-tuning, we update all the parameters of the network except the parameters of the convolutional feature encoder. For both, the BASE and the LARGE model, we used Adam optimiser with a learning rate of 1e-4 and a tri-stage learning rate schedule: the learning rate is warmed up for first 10\% of the steps, then held constant for the next 40\% steps and finally exponentially decayed for the remaining steps.
 
We set the maximum number of frames per GPU to 1M and fine-tune the models on 8 A100 GPUs without any gradient accumulation, making our effective batch size as 8M samples or 500 secs. We trained the BASE model for 80k steps and the LARGE model for 120k steps. For the first 200 steps, we only update the parameters in the final layer and then update all the parameters in the model (except those in the feature encoder). As a data augmentation, strategy, we mask the outputs of the feature encoder, similar to SpecAugment \cite{park2019specaugment}. We set the masking probability to 0.05 and the LayerDrop rate as 0.1 for both the models. 
For the other hyper-parameters, default values were taken from the wav2vec 2.0 code repository. Further we use early stopping with the patience set to 12 epochs. 
We used Fairseq \cite{ott2019fairseq} for our fine-tuning experiments.

\noindent\textbf{Language model setup} 
As mentioned earlier, we use a lexicon-based beam search decoder as implemented in Flashlight\footnote{https://github.com/flashlight/flashlight} library along with a language model to obtain predictions. For each language, we train 6-gram KenLM language model(s) \cite{heafield-2011-kenlm} with pruning for singletons of order 5 and doubletons of order 6. We consider three options for the corpus used for training the language model: (i) the IndicCorp corpus \cite{kakwani-etal-2020-indicnlpsuite} which contains 836 M, 1,860 M, 551 M, 719 M, 582 M, 674 M, and 107 M tokens in Bengali, Hindi, Marathi, Gujarati, Tamil, Telugu and Odia, respectively, (ii) the transcriptions available in the respective ASR datasets for each language, and (iii) a combination of the earlier two options. Similarly, we considered three choices for the lexicon: (i) the top 180K most frequent words from IndicCorp, (ii) the task-specific lexicon provided by the challenge organizers and (iii) combination of above two. Later, in Section~\ref{table:ablation-results}, we will show that the choice of lexicon and training corpus have a significant impact on the performance of the ASR system.

We tune hyperparameters $\alpha$ and $\beta$ in Equation \ref{eq:decoding} on the validation set, using grid search with values ranging from -4 to +4 and 0 to 5 for $\alpha$ and $\beta$, respectively. During this grid search, we use a beam size of 64. Note that $\alpha$ and $\beta$ are tuned independently for each language. Once the best value of $\alpha$ and $\beta$ are found, we use these for the test data and increase the beam size to 1024.  

\noindent\textbf{Rescoring setup} As mentioned earlier, we also consider using an external Transformer LM for rescoring the top-k hypotheses obtained from the decoder. 
For this, we train a BPE-tokenized Transformer LM using Fairseq toolkit\footnote{https://github.com/pytorch/fairseq} with 12 decoder layers, 16 attention heads, embedding dimension size of 1024 and FFN dimension size of 4096. 
The dataset used for training Transformer LM was exactly the same as that of KenLM with IndicCorp corpus included, while the vocabulary size was restricted to top 50k sub-words. 
The model was trained for 160k steps with the Adam optimizer and a tri-stage learning rate scheduler, comprising of 16k linearly increasing warm-up steps and a hold-up until 80k steps at learning-rate of 0.0001 after which the learning rate was decreased exponentially for the last 80k steps. 
Using this language model, we rescore the beams obtained from the decoder according to Equation \ref{eq:decoding2}. For selecting optimal values of $\alpha$ and $\beta$, we perform grid search in the range of 0 to 5 and -4 to +4 respectively.

\section{Results and Discussions}

\begin{table*}[]
\small\centering
\begin{tabular}{lcccccc|ccc}
\toprule
  & \multicolumn{6}{c|}{MUCS} & \multicolumn{3}{c}{OpenSLR} \\
 \cmidrule{2-10}
 & gu & hi & mr & or & ta & te & bn & ne & si \\
 \midrule
Best Models from Table \ref{table:ablation-results} & 14.3 & 9.5 & 11.7 & 17.2 & 19.5 & 15.1 & 10.5 & 9.4 & 18.6 \\
M4a: M4 + test lexicon (Oracle) & 14.5 & 10.8 & 8.6 & 16.5 & 15.5 & 14.9 & \textbf{8.0} & \textbf{5.6} & \textbf{12.1} \\
M5a: M5 + test lexicon (Oracle) & \textbf{13.0} & \textbf{8.6} & \textbf{7.5} & \textbf{14.8} & \textbf{14.2} & \textbf{11.3} & \textbf{8.0} & \textbf{5.6} & - \\
\bottomrule
\end{tabular}
\caption{Comparison of model performance when the test lexicon is used (oracle setting).}
\label{table:oracle-experiment}
\end{table*}

\begin{table}[]
\small
\centering
\begin{tabular}{lccc}
\toprule
 Fine-tuning data & \multicolumn{1}{c}{\textbf{ta}} & \multicolumn{1}{c}{\textbf{gu}} & \multicolumn{1}{c}{\textbf{te}} \\
 \midrule
1 Hour & 38.8 & 35.9 & 42.6 \\
$\quad$+LM$_{large}$+aug. lex. & 20.6 & 19.9 & 20.3 \\
10 Hours & 26.5 & 25.4 & 28.1 \\
$\quad$+LM$_{large}$+aug. lex. & 16.9 & 15.1 & 15.2 \\
20 Hours & 24.0 & 23.0 & 25.4 \\
$\quad$+LM$_{large}$+aug. lex. & 16.2 & 13.8 & 13.8 \\
40 Hour & 22.1 & 20.9 & 22.9 \\
$\quad$+LM$_{large}$+aug. lex. & 15.5 & 13.1 & 12.9 \\
\bottomrule
\end{tabular}
\caption{Comparison of model performance across varying amounts of fine-tuning data. We used a beam size of 128 keeping all other hyperparameters same as M6.}

\label{table:reduced-data-results}
\end{table}

\begin{table*}[h!]
\small\centering
\resizebox{\linewidth}{!}{
\begin{tabular}{lccc|cccccc|cc}
\toprule
 & \multicolumn{3}{c|}{MSR} & \multicolumn{6}{c|}{MUCS} & \multicolumn{2}{c}{OpenSLR} \\
 \cmidrule{2-12}
 & gu & ta & te & gu & hi & mr & or & ta & te & bn & ne \\
 \midrule
M6: & 12.3 & 15.1 &	12.4 & 14.8 &	10.5 & 12.2 & 21.9 & 20.0 & 15.2 & 10.6 & 9.7  \\
M8: $\quad$ + Multi. FT (Single Softmax) & 12.8 & 16.0 & 12.5 & 15.5 & 10.1 & 30.3 & 26.8 & 20.8 & 15.9 & 11.0 & 11.9 \\
M9: $\quad$ + Multi. FT (Multi-Softmax) \cite{joshi2021multiple} & 12.4 & 15.8 & 12.5 & 14.4 & 10.5 & 11.7 & 22.3 & 21.0 & 16.0 & 12.0 & 13.4 \\
\bottomrule
\end{tabular}
}
\caption{Comparison of multilingual fine-tuning strategies}
\label{table:multi-ft-results}
\end{table*}

\begin{table*}[h!]
\small\centering
\begin{tabular}{lccc|cccccc|cc}
\toprule
 & \multicolumn{3}{c|}{MSR} & \multicolumn{6}{c|}{MUCS} & \multicolumn{2}{c}{OpenSLR} \\
 \cmidrule{2-12}
 & gu & ta & te & gu & hi & mr & or & ta & te & bn & ne \\
 \midrule
M6: & 12.3 & 15.1 &	12.4 & 14.8 &	10.5 & 12.2 & 21.9 & 20.0 & 15.2 & 10.6 & 9.7  \\
M10: IPL with M6 checkpoint & 13.8	& 21.3 &	16.5 &	15.1 &	14.1 &	24.2 &	34.2 &	21.9 &	17.1 &	13.2 & 13.8 \\
\bottomrule
\end{tabular}
\caption{Comparison of different Iterative Pseudo-Labeling (IPL) strategies}
\label{table:ipl-results}
\end{table*}

\subsection{Ablation studies on pretraining}
We first discuss the effect of pretraining as well as various choices made during pretraining.

\noindent\textbf{No pretraining} In row 1 of Table \ref{table:ablation-results}, we present the results for an ASR model which has the same architecture as our BASE model but was trained from scratch (i.e., the weights were not initialised using pretraining). Comparing this to row M2, which is a model with the same architecture but is fine-tuned after pretraining, we see that pretraining significantly improves the performance. These results are consistent with similar results reported for English models in speech \cite{baevski2020wav2vec} as well as multilingual models in NLP \cite{DBLP:conf/nips/ConneauL19}. In a separate set of experiments (not included in Table \ref{table:ablation-results}) we found that pretraining is even more important when the model size is large. For example, training the LARGE model from scratch was highly unstable and even after 180K steps, the WER on the validation set was significantly higher as compared to a LARGE model which was initialised using pretraining (82.18 v/s 24.62).  

\noindent\textbf{Role of Pretraining Corpus} \citet{DBLP:journals/corr/abs-2107-07402} released a pretrained wave2vec model for 23 Indian languages using 9800 hours of speech data. In contrast, our model uses over 17,000 hours of pretraining data for 40 languages, with at least 500 hours of data in 15 languages. Comparing rows M2 and M3 of Table \ref{table:ablation-results}, we observe that our model trained on larger and more diverse data consistently improves over the model reported in  \citet{DBLP:journals/corr/abs-2107-07402}. In particular, we get an average improvement of 1.44 WER across all languages with a maximum improvement of 6.75 WER for Sinhala and a minimum improvement of 0.09 WER for Hindi. The improvements on Sinhala are noteworthy given that Sinhala is not present in the pretraining data of both the models. We believe that since our pretraining data has more diversity (40 languages as opposed to 23) and a better distribution of pretraining data across languages, it generalises better for languages not seen during pretraining. This is also evident given that our model performs better on Hindi despite a much smaller amount of Hindi pretraining data in our model (1075 v/s 4564 hours). These results establish importance of the larger pretraining dataset released as a part of this work.

\noindent\textbf{Effect of model size} Several works \cite{DBLP:conf/acl/ConneauKGCWGGOZ20,baevski2020wav2vec} have shown that large models pretrained on large amount of pretraining data outperform smaller models in downstream tasks. However, in the context of speech models for Indian languages, the right thresholds for data size and model size are not known. For example, with 17,000 hours of pretraining data would a LARGE model (317M parameters) outperform a BASE model (95M parameters). To answer this question we compare rows M3 and M4 of Table \ref{table:ablation-results} which only differ in the model size. We observe that the LARGE model consistently outperforms the BASE model with an average improvement of 2.08 WER across all languages, a maximum improvement of 7.22 WER for Marathi and a minimum improvement of 1.9 WER for Hindi. Again, a consistent reduction in WER across languages, including Hindi which has the largest representation establishes the value of larger model.

\noindent\textbf{Effect of model initialization}
As mentioned earlier, for pretraining our models we started with the pretrained English models released by \cite{baevski2020wav2vec}. This is in line with similar efforts in multilingual NLP where models pretrained for a larger set of languages are initialised with weights from a model pretrained for a smaller set of languages \cite{DBLP:journals/corr/abs-2008-00401}. To assess if this is needed for Indian speech models, we pretrained a LARGE model from scratch instead of initialising with the weights from the English model. We found this to be very expensive as even after 120k steps (which corresponds to 95 hours of training on 24 GPUs) this model had a loss of 2.17 on the validation data as opposed to the English-initialised model which had a loss of 1.82 after just 110k steps. This is an important finding as it suggests that as we expand to more low resource languages (beyond the 40 covered in our model), we should be able to leverage the pretrained checkpoints of our model and save costs on compute for training.

\subsection{Ablation studies on fine-tuning and decoding}
The previous section focused on discovering the right choices for pretraining. In this section, we focus on different choices for fine-tuning and decoding.

\noindent\textbf{Choice of language model} As mentioned earlier, language model (LM) and an accompanying lexicon play a crucial role in decoding the emissions of the acoustic model. Keeping the lexicon fixed (as extracted from the transcripts for the training and validation set), we consider two choices for the language model: (i) a LM trained using only the transcripts for the training and validation data, and (ii) a LM trained on much larger data from IndicCorp in addition to the transcripts for the training and validation data. Comparing rows M5 and M6 of Table \ref{table:ablation-results} we observe that integrating a LM trained on larger generic data outperforms a LM trained on smaller task specific data. We also observe the significant improvement in the WER by integrating a LM (using small data or large data) as compared to the WER of the acoustic model (row M4). Finally, rescoring the beams with an external Transformer LM, improves WER by an average of ~0.6 WER as shown by row M7 of Table \ref{table:ablation-results}.

To further investigate the role of language model in ASR, in Table \ref{table:reduced-data-results}, we experiment with different sizes of fine-tuning data (1 hour, 10 hours, 20 hours, 40 hours).  For the 1h dataset, the avg. WER came down from 39.1 WER to 20.3 WER, a reduction of 48.2\%, while for the 40h dataset, it reduced from 22.0 WER to 13.8 WER, a decrease of 21.4\%. This shows that the LM plays an important role when limited training data is available (which is typically the case for Indian languages). These observations lead to the following insight for developing ASR systems for low resource languages: \textit{build better LMs using (relatively) easily available raw text corpora in these languages}. 

\noindent\textbf{Effect of lexicon} 
Having established the importance of using LMs trained on larger data, we now fix the LM and change the lexicon. In particular, we augment the lexicon provided by task organizers with top 180K most frequent words from IndicCorp \cite{kakwani-etal-2020-indicnlpsuite}. The hope is that this augmented lexicon may reduce the number of OOVs during testing. Comparing rows M6 and M7 of Table \ref{table:ablation-results}, we observe that results are a bit mixed with a drop in WER for the MUCS and OpenSLR dataset but an increase in WER for the MSR dataset. One probable reason for worse WER in case of MSR dataset could be that the lexicon provided by task organisers for MSR challenge already contained tokens from test dataset and hence appending it with extra tokens degrades its quality. We believe that this is an error on the part of the organisers as test data should be excluded while building the lexicon (but we use it nonetheless to make our results comparable with other participants). To investigate this further, we did an oracle experiment where we further augmented the lexicon with words from the test set for the other datasets. We refer to this as an oracle experiment, because in practice we would not have access to the lexicon from the test set.

We present the oracle experiment results in row M4a, M5a of Table \ref{table:oracle-experiment}. With the new best models, we see an average improvement of 3.8 WER across languages, a maximum improvement of 6.5 WER in Sinhala and minimum improvement of 1.9 WER in Hindi.
We do not report these numbers in the Table \ref{table:ablation-results} as it would be unfair to compare this with the other results.
This leads to another insight: \textit{the right lexicon for the target domain can significantly help in low resource scenarios}. Of course, the lexicon should not be taken from the test set but independently curated. 

\noindent\textbf{Monolingual vs multilingual fine-tuning}
Some works \cite{DBLP:conf/acl/ConneauKGCWGGOZ20,joshi2021multiple} have shown that jointly fine-tuning a multilingual model using the task specific data in all languages leads to better performance. Motivated by these results, we jointly fine-tune the pretrained model by combining the training data from all the languages. 

During fine-tuning, we treat character units from each language as a unique token. We call this model as Single Softmax, as shown in row M8 of Table \ref{table:multi-ft-results} (since the characters across all the languages are competing with each other in a single softmax computation).
Comparing rows M6 and M8, which only differ in the fine-tuning strategy (monolingual v/s multilingual), we observe that the multilingual model gives a similar performance across all the languages. This is consistent with the results in multilingual NLP, where we typically see good gains due to a shared encoder (e.g. many-to-one translation) and less gains due to shared decoder (e.g. one-to-many translation) \cite{DBLP:journals/corr/abs-1907-05019}. One observation is that error rates on Marathi are significantly higher as for many sentences the model outputs Odia characters instead of Devanagari characters.  To address this issue, we follow \cite{joshi2021multiple} and use a multi-softmax layer, with each language having a separate softmax layer. The output vocabulary contains unique tokens from the respective language. During training, each mini-batch consists of data randomly sampled from a particular language. We show the results in row M9 of Table \ref{table:multi-ft-results}. We observe that the results for Marathi are improved significantly compared to row M8 while the performance remains consistent for other languages. These results are encouraging as they show that a single joint model (with a separate output layer for each language) gives performance which is comparable to that of individual models trained for specific languages. Such joint models are preferred as they are easier to maintain and deploy.

\noindent\textbf{Impact of fine-tuning dataset size}
A significant result reported in \citet{baevski2020wav2vec} was that pretraining a model using a large amount of unlabeled data significantly reduces the amount of fine-tuning data needed for building ASR models. In particular, they showed that starting with a pretrained model, using just 10 mins of fine-tuning data, led to single digit WERs for English. This is a significant result in the context of low resource languages. To check if the same holds true for Indian languages, we vary the amount of training data provided to the model from 1 to 10 to 40 hours. Quite disappointingly, our results as reported in Table \ref{table:reduced-data-results}, show that using just 1 hour of fine-tuning data leads to much worse performance. Further, we see gains in increasing the data from 20 hours to 40 hours, suggesting that adding more fine-tuning data would be beneficial. Lastly, note that we do not get a single digit WER for any of the languages compared to the results on English ASR in \cite{baevski2020wav2vec}. This indicates that developing ASR systems for Indic languages with their richer phoneme/character sets and vocabulary is more challenging than English.

\begin{table}[]
\small
\resizebox{\linewidth}{!}{
\begin{tabular}{lccccccc}
\toprule
 & hi & gu & mr & or & ta & te \\
 \midrule
Baseline & 27.45 & 25.98 & 20.41 & 31.28 & 35.82 & 29.35\\ \midrule
CSTR & \underline{14.33} & 20.59 & 15.79 & 25.34 & 23.16 & 21.88 \\
BSA & 16.59 & 21.30 & 15.65 & 17.81 & 28.59 & 25.37 \\ 
EM & 17.54 & \underline{20.11} & 20.15 & 19.99 & 28.52 & 26.08 \\
EkStep & 12.24 & 30.65 & 39.74 & 27.10 & 27.20 & 22.43 \\
Uniphore & 22.79 & 22.79 & \underline{14.9} & 29.55 & \textbf{\underline{18.8}} & 28.69 \\
Lottery & 17.81 & 23.62 & 58.78 & \underline{17.74} & 30.69 & 27.67 \\
IIITH & 31.11 & 26.94 & 33.8 & 37.19 & 35.03 & \underline{17.00} \\ \midrule
M5: & 14.7 & 17.2 & 13.8 & \textbf{17.2} & 25.0 & 20.5 \\
M6: & \textbf{10.5} &	\textbf{14.8} & \textbf{12.2} & 21.9 & 20.0 & \textbf{15.2} \\

\bottomrule
\end{tabular}
}
\caption{Comparison of our best models (M5, M6) with the the top performers from the MUCS 2021 leaderboard. Individual best models from the leaderboard are underlined.}
\label{table:mucs-results}
\end{table}

\begin{table}[h!]
\centering
\small
\resizebox{\linewidth}{!}{
\begin{tabular}{lccc}
\toprule
 & gu & ta & te \\
 \midrule
\begin{tabular}[c]{@{}l@{}}Baseline \cite{srivastava18_sltu} \end{tabular} & 19.8 & 19.4 & 22.6 \\ \midrule
Jilebi \cite{FITPUB11841} & 14.0 & 13.9 & 14.7 \\
Cogknit \cite{DBLP:conf/interspeech/FathimaPCI18} & 17.7 & 16.0 & 17.1 \\
\begin{tabular}[c]{@{}l@{}}CSALT-LEAP\\ \cite{srivastava18_sltu} \end{tabular} & - & 16.3 & 17.6 \\
ISI-Billa \cite{Billa2018ISIAS} & 19.3 &  19.6 & 20.9 \\
MTL-SOL \cite{Sailor2020MultilingualSR} & 18.4 & 16.3  & 18.6 \\
Reed \cite{9383457} & 16.1 & 19.9 & 20.2 \\
\begin{tabular}[c]{@{}l@{}}CNN + Context temporal features\\ \cite{Sen2020ShortTC} \end{tabular} & 18.4 & 24.3 & 25.2 \\
EkStep model$^{*}$ & 19.5 & 22.1 & 21.9 \\ \midrule
M5: & \textbf{11.7} & \textbf{13.6} & \textbf{11.0} \\
M6: & 12.3 & 15.1 & 12.4 \\
\bottomrule
\end{tabular}
}
\caption{Comparison of our best models (M5, M6) with the the top performers from the MSR 2018 leaderboard as well as other recent state of the art methods. }
\label{table:msr-results}
\end{table}

\begin{table}[]
\small
\resizebox{\linewidth}{!}{
\begin{tabular}{lccc}
\toprule
 & bn & ne & si \\
 \midrule
Baseline \cite{9414961} & 17.9 & 12.9 & 21.8 \\ \midrule
Ekstep model$^{*}$ & 15.2 & 13.8 & 20.0  \\ \midrule
M4: & 14.3 & 13.0 & \textbf{18.6} \\
M6: & \textbf{10.6} & \textbf{9.7} & - \\
\bottomrule
\end{tabular}
}
\caption{Comparison of our best models (M4, M6) with state-of-the-art results reported in the literature. * The Ekstep model was fine-tuned by us.}
\label{table:openslr-results}
\end{table}

\subsection{Iterative Pseudo Labelling}
We also explore augmenting the labelled datasets with Iterative Pseudo-Labeling (IPL) \cite{xu2020iterative} which is a semi-supervised algorithm for end-to-end ASR applied on larger unlabelled datasets. 
Motivated by the success of this approach \cite{xu2020iterative,baevski2020wav2vec}, we explore the use of IPL on our large unlabelled pretraining corpus.
We randomly sample 500 hours of pretraining data for each language. Then, using our monolingual models (M6) we generate pseudo-labels for the data. Similar to the fine-tuning setup, we use a lexicon-based beam search decoder using LM$_{large}$ + augmented lexicon with beam size of 128. To eliminate noise from the data, we remove transcriptions having less than 3 words. We observe that $\sim$10\% of the transcriptions are filtered out. Next, we perform a second round of fine-tuning combining the pseudo-labeled data along with the fine-tuning datasets. 
We present the results of iterative pseudo-labeling in Table \ref{table:ipl-results}. We initialize the models from the fine-tuned checkpoints (M6) and perform IPL, which is shown in row M10.
We observe that the best performance is retained by models that do not use IPL, \textit{i.e.}, M6. 
This is a surprising finding as IPL has been shown to work very well for English \cite{xu2020iterative}. We believe that this is not the case for Indian languages because the performance of the Indian models used for pseudo labeling the data is poorer as compared to English ASR models (WERs $>10$ as opposed to $<5$ for English). Thus, the generated pseudo-labels would be more noisy in our case. Furthermore, the unlabeled audio data that we used does not contain sentence-wise segments unlike the LibriSpeech dataset \cite{panayotov2015librispeech}. This causes the transcriptions to be noisy at the boundaries of the audio sample. 
These issues need to be resolved before establishing the value of IPL for Indic ASR.

\subsection{Comparison with existing baselines}
We now compare with existing models on the three datasets. For each dataset, we compare with the following models:

\bulletc \textbf{Baseline.} This is the baseline model as reported by authors of the dataset (OpenSLR) or the organisers of the challenge (MUCS/MSR). Often, this is a Time delay neural network (TDNN) model \cite{DBLP:conf/interspeech/PeddintiPK15}.

\bulletc \textbf{Top3 entries from the leaderboard.} For the MUCS\footnote{https://navana-tech.github.io/MUCS2021/leaderboard.html} and MSR\footnote{ https://www.microsoft.com/en-us/research/event/interspeech-2018-special-session-low-resource-speech-recognition-challenge-indian-languages/\#!leaderboard} challenge, we include results for the Top 3 entries from the leaderboard of the challenge. Note that the MUCS challenge has only recently concluded and the system papers written by the participating teams are not yet available. 
\bulletc \textbf{Best per language.} For the MUCS challenge, we observed that some systems did  poorly when considering the average performance across all languages, but did very well on a specific language (and hence did not appear in the top3 entries on the leaderboard). For example, one participant, Uniphore, got an average WER of 22.92 ranking 4th on the leaderboard but had the best performance for Tamil with a WER of 18.8. For a comprehensive comparison, for every language we report the \textit{best individual system} for that language even if its performance was very poor on other languages.

\bulletc \textbf{Other SOTA systems.} Given that the MSR challenge was launched in 2018, there are some follow-up works \cite{Sailor2020MultilingualSR,9383457,Sen2020ShortTC} which have reported results on this dataset. We report the numbers as it is from these works. Similarly, for OpenSLR, we compare with the results reported in \citet{9414961}.

\bulletc \textbf{EkStep.} As mentioned earlier, this is the recently released pretrained model covering 23 Indian languages \cite{DBLP:journals/corr/abs-2107-07402}.

The above results are summarised in Table \ref{table:mucs-results}, \ref{table:msr-results}, \ref{table:openslr-results}. Across all the datasets and all the languages (except MUCS-Tamil), we establish new state of the art results. Even for MUCS-Tamil, we outperform the top3 models on the leaderboard.
These results clearly establish the recipe for developing ASR systems for Indian languages: \textit{multilingual pretraining followed by multilingual fine-tuning combined with language models trained on large corpora with good lexicons.}

\section{Analysis of the pretrained model}
IndicWav2Vec has been trained on 40 languages and represents the largest diversity of Indian languages in any such multilingual model.
In this section, we analyze this pretrained model to understand (a) the multilingual representations learnt by the model, and (b) the role of the attention heads that are characteristic of Transformer layers in the model.

\subsubsection{Multilingual representations}

\begin{figure}[t]
\centering
\includegraphics[width=\columnwidth]{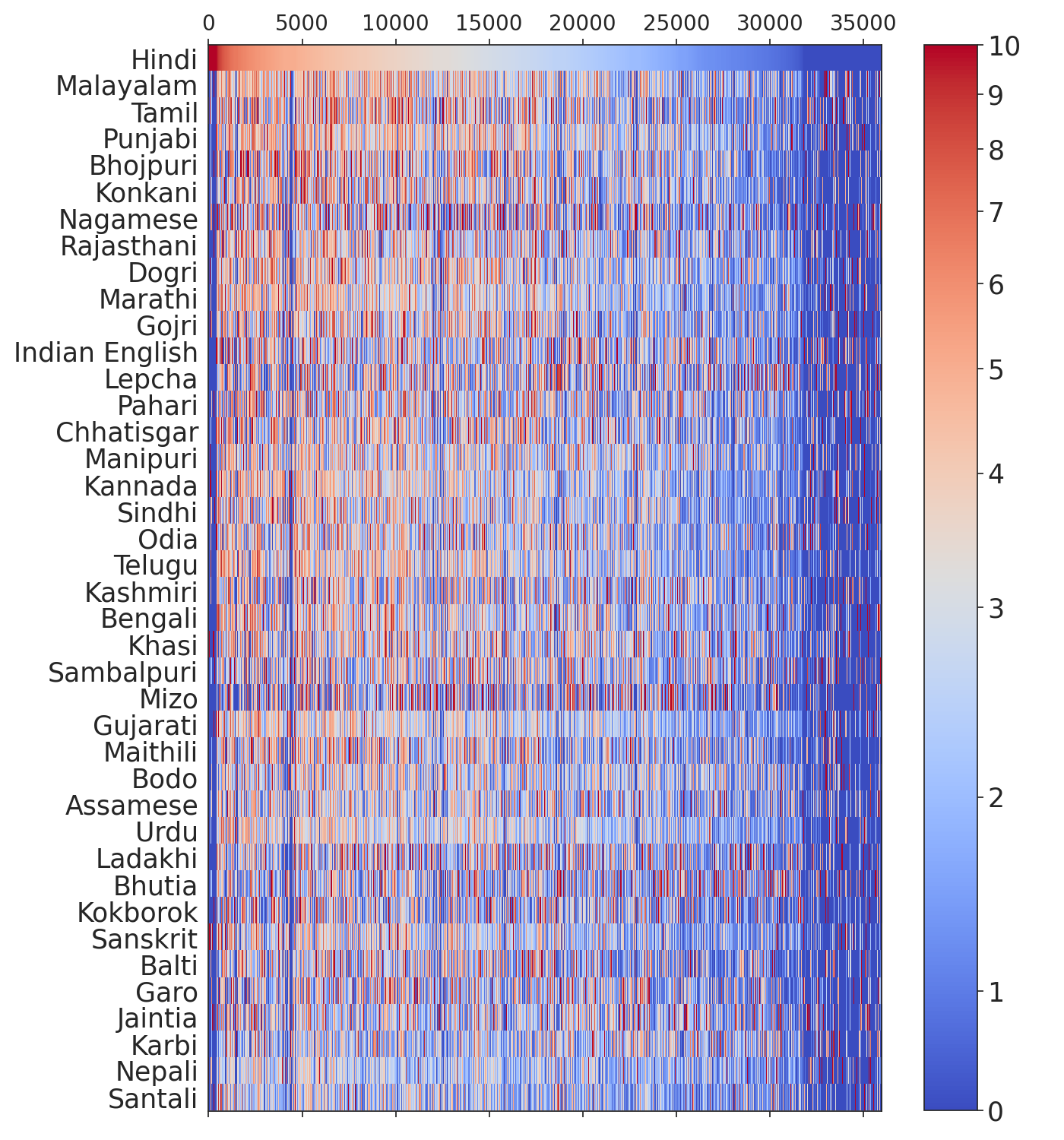}
\caption{The plot shows the distribution of codebook entries across languages.}
\label{fig:analysis:codebookheatmap}
\end{figure}

As discussed, the first part of the IndicWav2Vec model computes representations of frames of audio from a learnt codebook.
In the model, we use a codebook of 102.4K (320 $\times$ 320) vectors which are shared by all 40 languages.
One may ask two questions about these code-books: (a) Are codebook entries disjointed or common across languages?, and (b) Is the sharing of codebook entries stronger between more related languages?
To answer this, we visualize the codebook entries (columns) across all languages (rows) in the heatmap shown in \ref{fig:analysis:codebookheatmap}. 
We randomly sample 5 hours of training data samples from each language, i.e. a total of 200 hours of data. We generate codebook vectors by passing the audio sequence through the feature encoder followed by the quantizer. The color of each cell denotes the frequency of that code-book entry for that language - wherein red denotes higher frequency and blue lower-frequencies.
Here the columns are sorted based on decreasing order of frequency of occurrence of code-book entries in Hindi audio.
The rows are sorted in decreasing order of similarity of the frequency pattern of that language with Hindi.
As can be seen from the first few columns which have red cells in many rows, a large number of the code-book entries common in Hindi are also used in other languages. 
This indicates the value in multilingual pretraining. 
However, it is also clear that as we move to languages of other families, such as Mizo, there are large numbers of codebook entries that are unique to these languages.
For instance, a very few of the early red columns for Hindi have similarly high frequency counts for Mizo. 

\begin{figure}[t]
\centering
\includegraphics[width=\columnwidth]{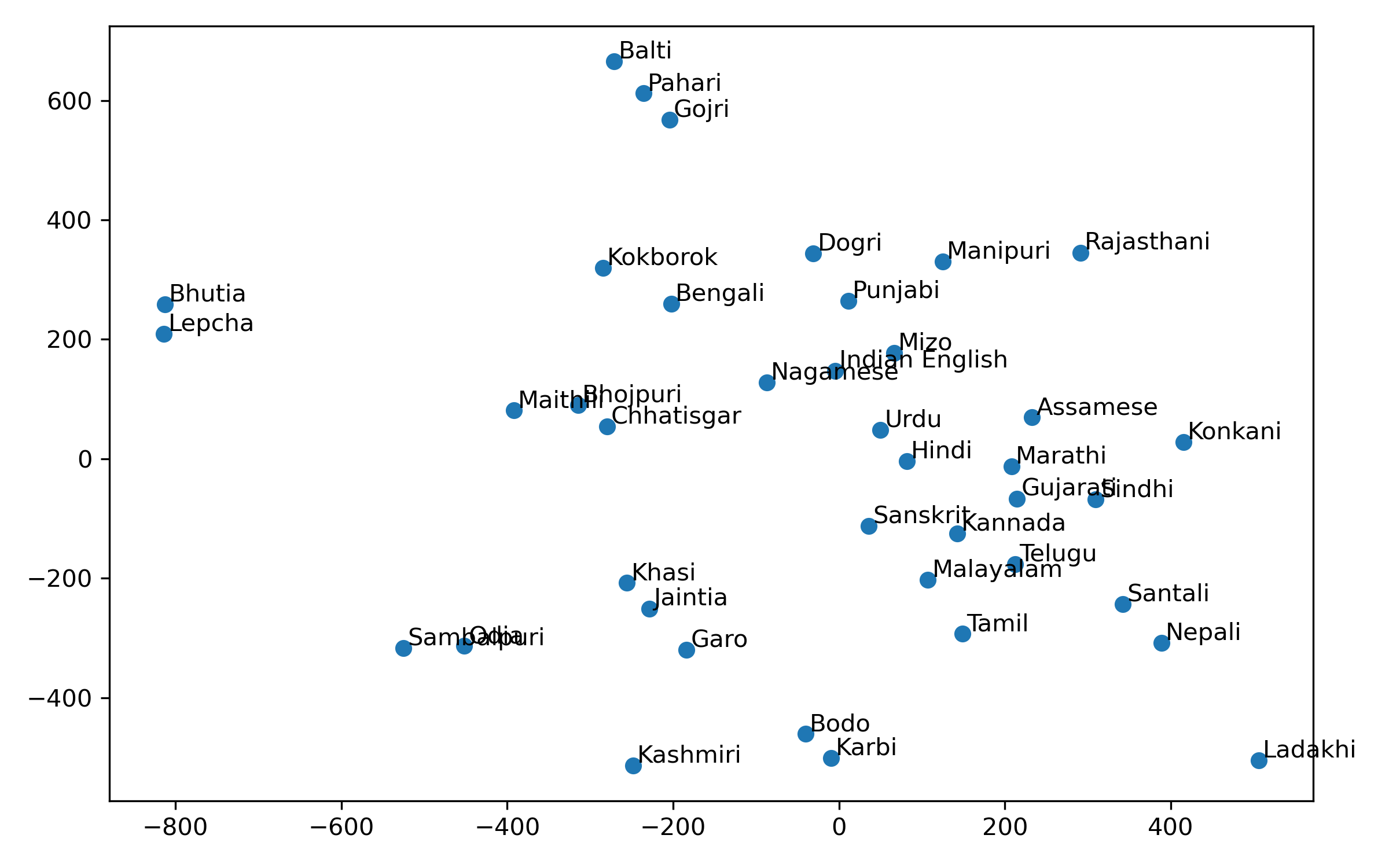}
\caption{t-SNE plot showing language representations learnt at the 5-th layer of the context network.}
\label{fig:analysis:centroids}
\end{figure}

While the above visualization is from the codebook entries, we may also study the representations learnt by the model across the layers of the Transformer.
We use the same 200 hours of sampled data, and generate the transformer layer outputs for each sequence.We take an average of all frames in a sequence to compute a single representation. Next, we compute the centroids of all representations for a given language.
We plot the centroids language-wise after a t-SNE projection in \ref{fig:analysis:centroids}.
As the plots shows several related languages (such as Kannada and Telugu; Hindi and Urdu; Khasi, Garo and Jaintia; Bodo and Karbi;  Balti, Pahari and Gojri; Bhojpuri and Maithili) are close to each other, while some languages (Bhutia and Lepcha) are quite distinct from others.
This plot again reaffirms that multilingual pretraining has the potential to generalize representation learning across similar languages.

\subsubsection{Role of Attention Heads in Transformer}

\begin{figure}[t]
\centering
\includegraphics[width=\columnwidth]{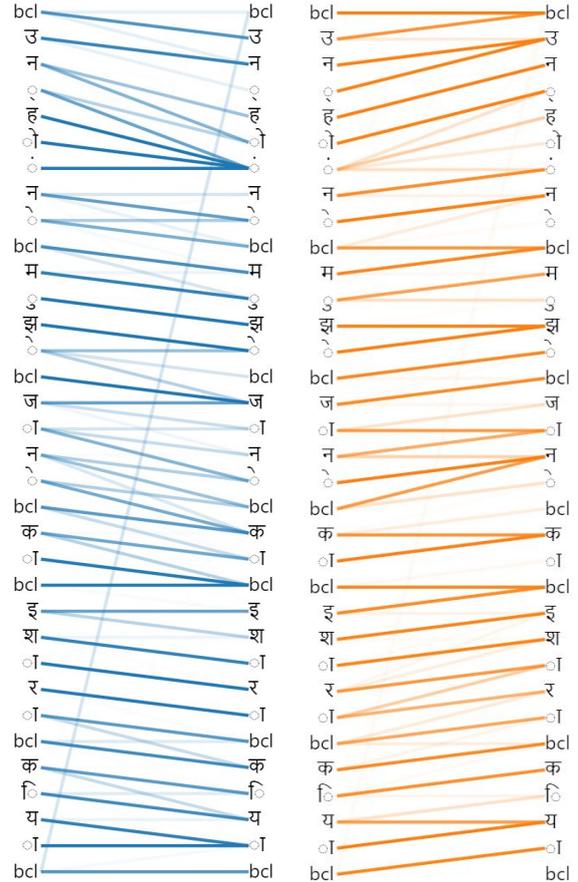} 
\caption{Attention heads pay attention to next phoneme, previous phoneme, and silence tokens. 'bcl' indicates the silence token.}
\label{fig:analysis:attention}
\end{figure}

The attention heads which characterize Transformers are crucial for sharing information between the representations of multiple frames in a sequence. 
In the NLP literature, extensive studies have been performed to analyze and interpret the role of these attention heads \cite{pande2021heads}.
To the best of our knowledge, no similar study exists for Transformer architectures for ASR.
For an attention head in a Transformer layer, let $a_{ij}$ denote the attention paid in updating the representation of frame $i$ using the representation of frame $j$. 
We make use of our model fine-tuned on the Hindi MUCS dataset to generate phoneme annotations for each frame. We use the predictions computed by the CTC \cite{10.1145/1143844.1143891} loss to extract this information. Given that frames are relatively short and a phoneme spans multiple frames, we compute an aggregation of $a$ given as $\tilde{a}_{pq} = \sum_{i \in P} \sum_{j \in Q} a_{ij}$ where $p$ and $q$ are phonemes and $P$ and $Q$ are the frames respectively that comprise the two phonemes. 
We compute $\tilde{a}$ for pairs of phonemes for a few sample input sentences and plot them in \ref{fig:analysis:attention}.
We make two key observations from these plots.
First, attention heads seem to play some distinctive roles - a major fraction of them either attend to nearby phonemes (next or previous) or to the nearest silence.
This is similar to findings in NLP where a majority of the attention heads encode local and a few heads encode syntactic information \cite{pande2021heads}.
Second, this clear pattern in the attention heads motivates building efficient models that regularize attention heads to attend to neighboring phonemes and silence. 
We note this a potential future work.

\section{Discussion and Recommendations}
Based on our work, we make the following recommendations for developing ASR systems for Indian languages.
\noindent\textbf{Informing ASR model choice}
Our results comparing IndicWav2Vec$_{large}$ models with several model architectures proposed over the last several years suggests that a Wav2Vec like model achieves state-of-the-art results as also evidenced in English ASR.
We also establish that larger models which can be pretrained with larger corpora of multilingual data are beneficial.

\noindent\textbf{Informing ASR metrics for evaluation}
In Table \ref{table:ablation-results}, we observed that the Language Model and lexicon play a significant role in improving the WER. We further report experiments which show that, the large gap in the WER of two acoustic models, significantly reduces when the same LM is integrated with both the models. Hence, when evaluating  architectural choices made by different acoustic models (such as a better feature encoder, better quantizer, better vocabulary, etc) it is important to ensure that the LMs and lexicons used by the models being compared are exactly the same.  

\noindent\textbf{Informing ASR benchmark creation} Given the role of LM and lexicon, it is important to carefully design benchmarks for evaluating ASR systems. In particular, the distinction between performance on test sets with a large number of OOVs v/s test sets with mainly in-domain vocabulary should be clearly made. This requires the creation of benchmarks which contain data from different domains allowing a more nuanced evaluation of ASR models.

\noindent\textbf{Informing ASR resource creation} The finding from English ASR that fine-tuning can be performed with just tens of minutes of labelled data does not transfer to Indic languages.
In particular, we find that both pretraining and fine-tuning datasets sensitively affect accuracy.
Thus, model or training choices do not seem to substitute the continued need for collecting resources for Indic languages.

\noindent\textbf{Extending to the long tail of languages} Our results show that pretraining with a diverse set of languages, also improves the performance on related languages which our not present in the pretraining data (e.g., Sinhala). This creates hope for developing ASR systems for the long tail of languages in India. However, to enable this, we still need efforts to create evaluation benchmarks and fine-tuning data (at least 40 hours) for these languages. This is a call to the community to support us in this effort of developing ASR systems for the next billion users. 

\section{Conclusion}
We report results of applying two recent and successful ideas from English ASR to Indic ASR: use of wav2vec like model architecture and use of unlabelled data to pretrain the model.
We implement this with a curated dataset on Indic languages and a range of ablation studies on architecture, pretraining, fine-tuning, and decoding choices.
Through this, we obtain state-of-the-art results on 9 Indic languages across 3 datasets.
While advancing ASR systems for the next billion users from the sub-continent, our results highlight the need for larger resources and benchmarks across more languages.

All the models developed as a part of this work, \textit{viz.}, the pretrained model, the language-specific fine-tuned models and the language models along with the Fairseq and KenLM scripts and configuration files used for building them will be publicly released. We hope that these models will help in advancing the state of the art for Indian Speech Technology.

\section*{Acknowledgments}
We would like to thank the Ministry of Electronics and Information Technology (MeitY\footnote{https://www.meity.gov.in/}) of the Government of India and the Centre for Development of Advanced Computing (C-DAC\footnote{https://www.cdac.in/index.aspx?id=pune}), Pune for generously supporting this work and providing us access to multiple GPU nodes on the Param Siddhi Supercomputer. We would like to thank the EkStep Foundation for their generous grant which went into hiring human resources as well as cloud resources needed for this work. We would also like to thank the Robert Bosch Center for Data Science and Artificial Intelligence for supporting Sumanth and Gowtham through their Post Baccalaureate Fellowship Program.

\bibliography{anthology,custom}
\bibliographystyle{acl_natbib}




\end{document}